\pdfoutput=1
\documentclass[11pt]{article}

\usepackage[final]{acl}

\usepackage{times}
\usepackage{latexsym}

\usepackage[T1]{fontenc}
\usepackage[utf8]{inputenc}

\usepackage[most]{tcolorbox}
\usepackage{xcolor}

\definecolor{lightgray}{gray}{0.9}
\definecolor{boxblue}{RGB}{220, 230, 241}
\definecolor{boxyellow}{RGB}{255, 244, 204}
\definecolor{lightgreen}{RGB}{220, 250, 220}

\usepackage{microtype}

\usepackage{inconsolata}

\usepackage{graphicx}

\title{Large Language Models and Provenance Metadata \\ for Determining the Relevance of Images and Videos in News Stories}

\author{Tomas Peterka \\
  Gymnazium Jana Keplera \\
  \texttt{xpetto01@gjk.cz} \\\And
  Matyas Bohacek \\
  Stanford University \\
  \texttt{maty@stanford.edu} \\}

\begin{document}
\maketitle
\begin{abstract}
The most effective misinformation campaigns are multimodal, often combining text with images and videos taken out of context---or fabricating them entirely---to support a given narrative. Contemporary methods for detecting misinformation, whether in deepfakes or text articles, often miss the interplay between multiple modalities. Built around a large language model, the system proposed in this paper addresses these challenges. It analyzes both the article's text and the provenance metadata of included images and videos to determine whether they are relevant. We open-source the system prototype and interactive web interface.
\end{abstract}

\section{Introduction}

"Hospital in NYC full of dead bodies smfh," reads a tweet from April 3, 2020, in the early days of the COVID-19 pandemic~\cite{newslit}. Attached is an image showing body bags lined up on a hospital floor. The tweet quickly went viral. But the photo was taken out of context---it had been captured years earlier in Ecuador. Despite this, it fueled the spread and engagement of this false narrative.

Leveraging out-of-context media to strengthen a false narrative is a popular technique in online misinformation. Last year, it was used by over $40 \%$ of online misinformation~\cite{dufour2024ammeba}. But mere usage of imagery that is not relevant is not the only technique through which visual media can advance inaccurate information. From basic image and video editing, accounting for over $40 \%$ of content-based misinformation, to AI-generated image and video, accounting for over $30 \%$ of content-based misinformation, media-powered misinformation is on the rise, overtaking pure text-based misinformation~\cite{dufour2024ammeba}.

Unlike video deepfakes showing a prominent figure (such as a politician or a celebrity) delivering a manipulated speech, which can be detected from the video signal in isolation~\cite{farid2022creating}, and false AI-generated images, which can be detected from visual artifacts~\cite{cozzolino2024raising}, misinformation leveraging out-of-context media is often more challenging to detect~\cite{papadopoulos2024similarity,abdelnabi2022open}. One of the main reasons is that such detection requires a multi-modal analysis reflecting the semantics of the narrative and the media. While there have been some methods striving to tackle this task in the literature, these often rely on supervised models that are topic-specific and struggle to generalize to future data~\cite{agarwal2019analysis,aimeur2023fake}.

In light of these challenges, data provenance recently emerged as a promising avenue for establishing key events in lifecycle of digital media~\cite{england2021amp,feng2023examining}, such as when, how, and where it was created, along with any subsequent modifications~\cite{sherman2021designing}. While approaches to data provenance differ in the extent of information that is recorded in the metadata and the level of protection against manipulation, they typically include details about the method of origin (e.g., captured on a camera, generated using an AI model), time and place of origin, and any edits (e.g., actions in Photoshop, AI edits). Provenance metadata can also include information about the author and owners over time. Often, this metadata is protected against tampering through a central issuing entity or blockchain.

We propose a method that leverages provenance metadata to tackle the problem outlined above---determining whether a given media is relevant to the news article. Built on top of a large language model (LLM), our method is presented with the text of a news story, a visual caption of the attached media, and the provenance metadata of the attached media. The method then determines if the attached media fit the context and whether they have not been tampered with. This analysis is accompanied by reasoning; the LLM can even be prompted with follow-up questions. Beyond describing this method, we present a prototype implementing the method with a specific data provenance framework and a specific LLM, presented as a protoype web interface. The code of this prototype is open-sourced\footnote{\url{https://github.com/matyasbohacek/media-relevance-in-news-stories}} under the MIT license.

We open this paper by briefly reviewing prior work on this topic. We then describe our system and the prototype web interface, as well as the implications and limitations of this work. Finally, we outline productive questions for future work. 

Our discussion and evaluation scope is limited to news stories as we seek to provide a focused discussion of our method. However, we expect that it can be applied to social media, blog posts, and other contexts with minimal adaptation.

\begin{figure*}[t] % 'p' for a full page
    \centering
    \begin{tabular}{cc}
        \includegraphics[width=0.48\textwidth]{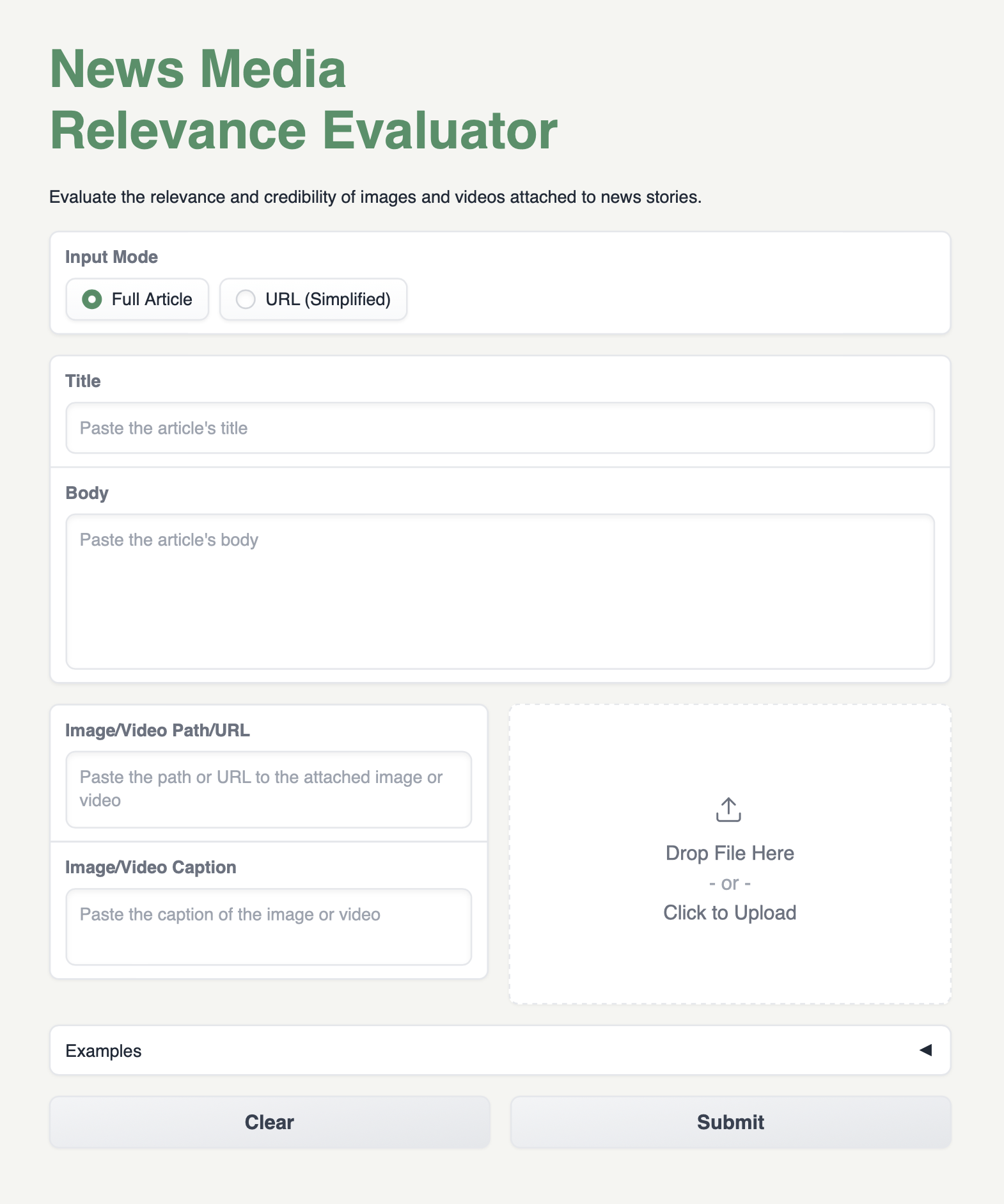} &
        \includegraphics[width=0.48\textwidth]{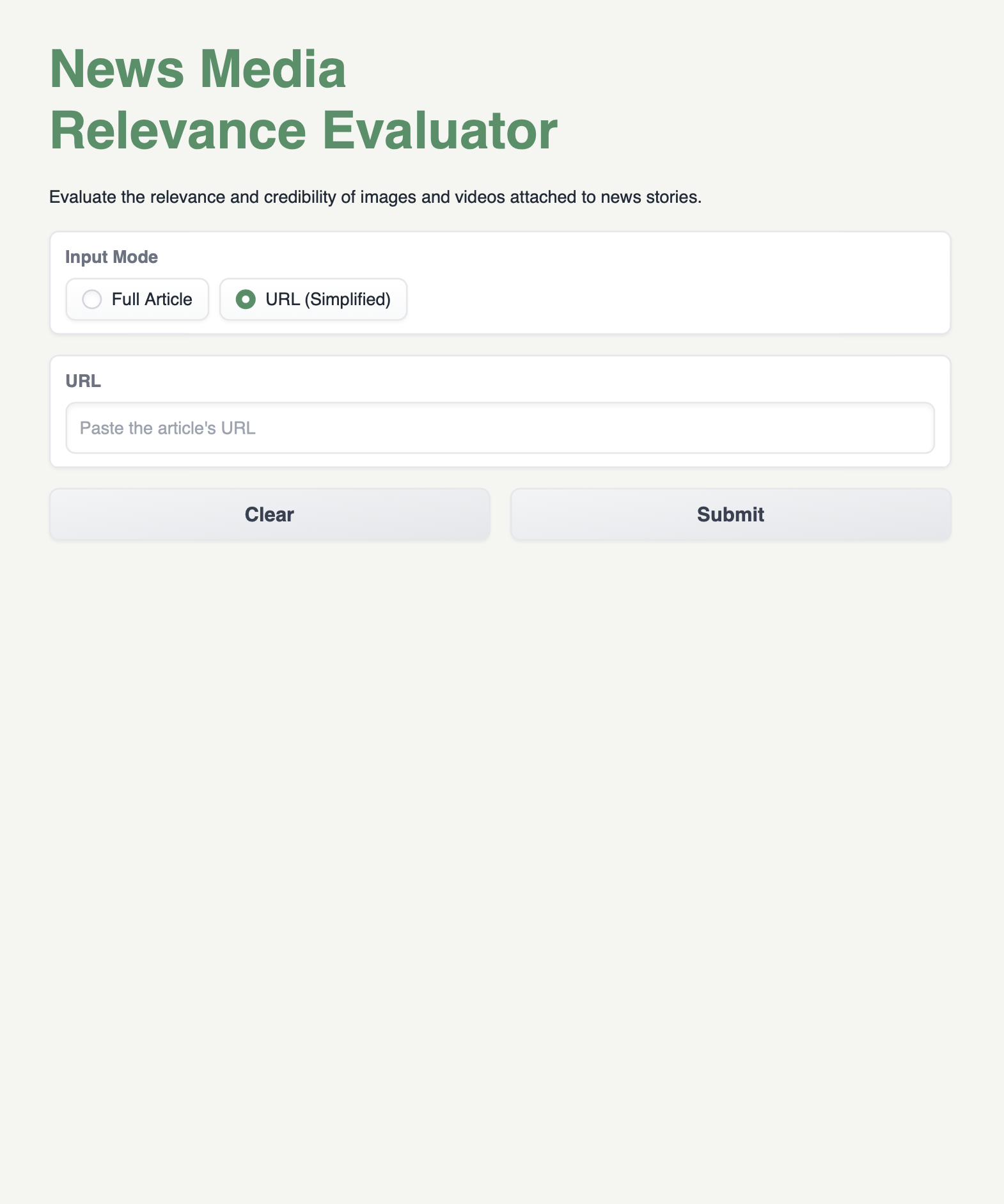} \\
        (a) & (b)
    \end{tabular}
    \caption{Screenshots of the (a) full article and (b) URL (simplified) input interface in the prototype web interface.}
    \label{fig:input-interface}
\end{figure*}

\section{Related Work}

We briefly review existing literature tackling the relevance of visual media in misinformation. Due to the short format of this paper, this review is limited, and we refer the reader to \citet{micallef2022cross}, \citet{bu2023combating}, and \citet{hartwig2024landscape} for a more extensive review.

It has been shown that the presence of visual media in misinformation affects the attention, comprehension, memory, and judgment of its audience~\cite{newman2023misinformed}. In addition to the effects on readers, the presence of visual media also increases the chances of such narratives being amplified by the recommendation algorithms on social media and in search engines~\cite{li2020picture,abell2023digital}. Put together, it comes as no surprise that the use of visual media in misinformation is rapidly increasing~\cite{yang2023visual, dufour2024ammeba}.

One of the first methods aiming to tackle this problem was a supervised model for the classification of image relevance~\cite{aneja2021cosmos}. However, this method only considers the image and caption pair, leaving out the rest of the article.

Later work proposed to use reverse image search to address this problem~\cite{qian2023fighting}, seeking to empower users to proactively verify the origin of images in articles. When tested in a participant study, however, \citet{qian2023fighting} found mixed results of efficacy.

Recent work focused on employing automated methods analyzing the semantic coherence of various modalities in a single article~\cite{xu2024mmooc}. While generally effective, this approach misses critical nuance in cases where the use of older images or images from different places than the main story may be an appropriate practice.

Finally, a recent approach introduced LLMs to the task by fine-tuning a multimodal LLM~\cite{qi2024sniffer}. While leading to effective explanations, this approach does not account for the possibility of media being visually relevant to the article but actually taken at a place or time that is not. Similarly, it does not account for tampering with the media.

\begin{figure*}[t] % 'p' for a full page
    \centering
    \begin{tabular}{cc}
        \includegraphics[width=0.48\textwidth]{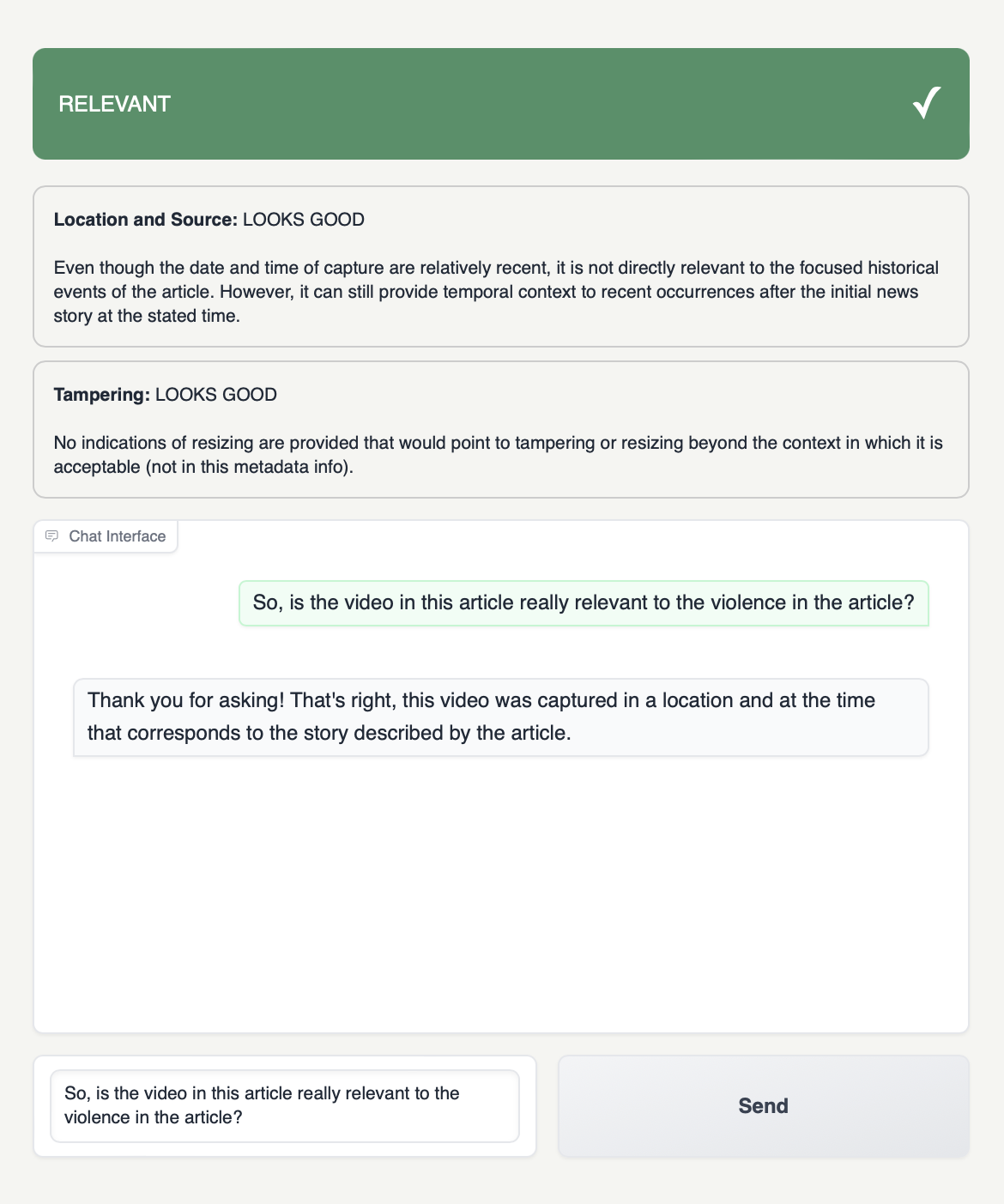} &
        \includegraphics[width=0.48\textwidth]{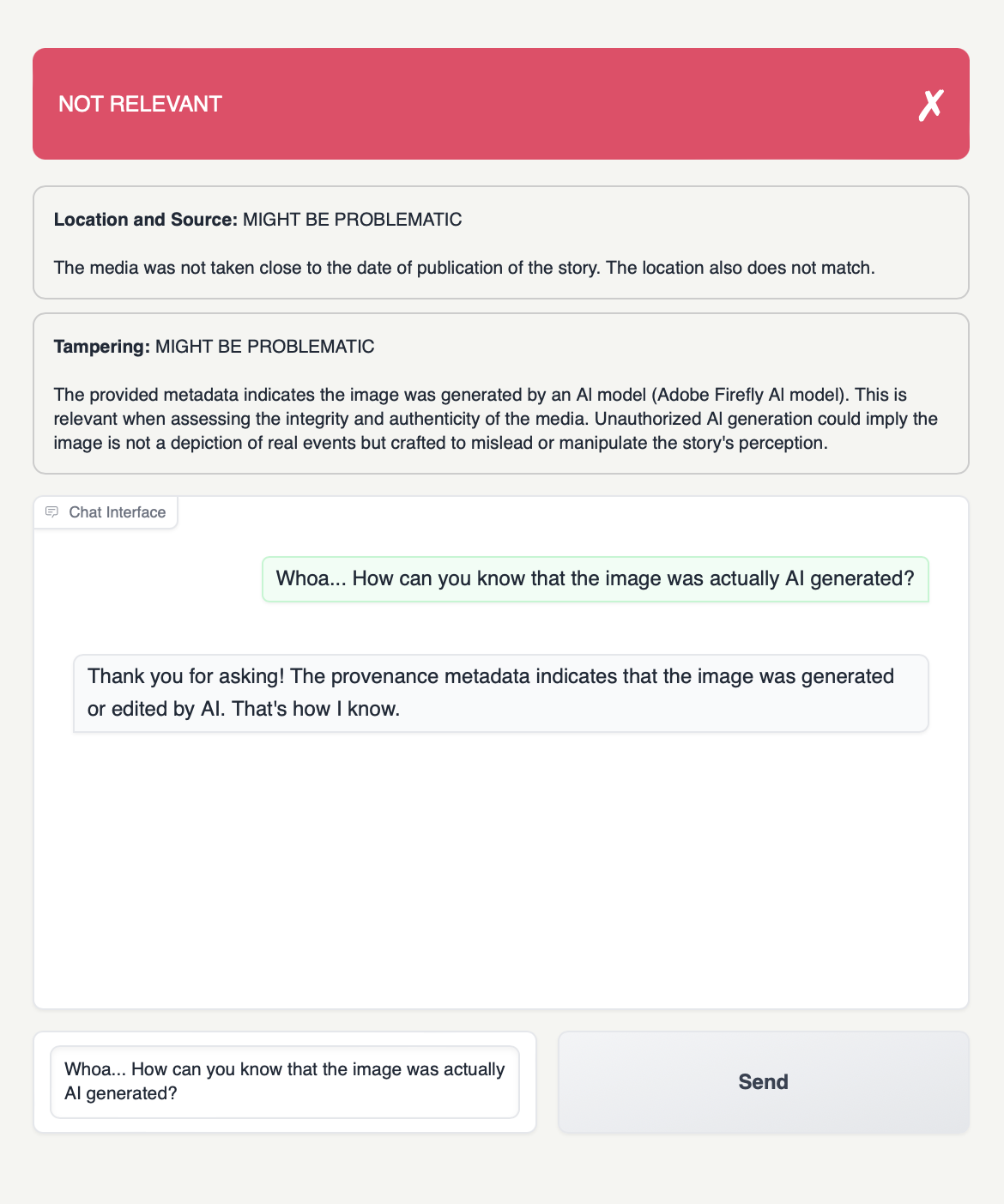} \\
        (a) & (b)
    \end{tabular}
    \caption{
    Screenshots of the prototype web interface displaying a result in which the media were found (a) relevant and (b) not relevant to the news story. The chat interface allows for submitting follow-up questions to the LLM.
    }
    \label{fig:result}
\end{figure*}

\section{Proposed Method}
\label{sec:method}

We present a method for the following task: given a news article, determine whether the included media (e.g., images and videos) are relevant to it. By relevant, we refer to media that were captured at the location and time of the reported event or that present content that is relevant to the story as generic illustrative material.

Our method uses data provenance to establish facts about the media's date and location of origin and LLMs to determine whether the origin is relevant to the body of the news article. The motivation for using provenance metadata, especially when the implemented framework offers cryptographic guarantees, is the accuracy and authenticity of this information. The motivation for using LLMs is to leverage their out-of-the box knowledge, without the need for training or fine-tuning.

As input, the method is presented with an article consisting of a title, body, and attached media (images, video, or audio). The output comes an analysis determining whether (a) the origin of the media is relevant, (b) the edits made to the media are relevant, and (c) whether it is overall relevant.

The specific steps taken by the method include:

\begin{enumerate}
    \item (\textit{Optional}) Structured article data (title, body, media) is scraped from the article's URL. This step is skipped if the structured article data is explicitly provided.
    \item The provenance metadata of any attached media (e.g., images and videos) are extracted and converted to an LLM-readable format.
    \item The article and the provenance metadata are presented to the LLM for analysis.
    \item The LLM returns its analysis, which is filtered and structured into an overall assessment.
    \item (\textit{Optional}) The LLM can respond to follow-up questions about the analysis.
\end{enumerate}

Note that the proposed method can, by design, work with different data provenance frameworks and LLMs, and hence leverage improved capabilities of both technologies as they continue to evolve.

\section{Prototype and Web Interface}
\label{sec:prototype}

In this section, we describe a specific prototype implementation to demonstrate the method proposed in Section~\ref{sec:method}. We first provide an overview of the specific building blocks used for the prototype, and then describe the web interface built to interface this prototype.

\subsection{Article Scraping}

Structured article data is extracted from the article's HTML using the Newspaper4k library~\cite{newspaper4k}. In particular, this library extracts the title, body, and attached media (images, video, or audio) based on standard HTML conventions and heuristics.

\subsection{Provenance Metadata}

The C2PA standard~\cite{rosenthol2022c2pa} is used to extract the provenance metadata. Fields relevant to the location and time of origin and modifications made to the image are filtered and converted into an LLM-readable format (see our implementation\footnote{\url{https://github.com/matyasbohacek/media-relevance-in-news-stories}} for details).

\subsection{Large Language Model}

The analysis is performed using the Phi-3 LLM~\cite{abdin2024phi}, which was used for its small size and efficacy. The system and query prompts are included in Appendix~\ref{app:prompt}. It was implemented using the Hugging Face Transformers libraries~\cite{wolf2019huggingface}.

For the purposes of the prototype web interface and to induce a chain-of-thought behavior of the LLM, the prompt asks the model to provide reasoning for the assessment. This reasoning is not a required part of the proposed method but rather a mechanism for identifying the model's weak spots.

\subsection{Web Interface}

The web interface is implemented in Gradio~\cite{abid2019gradio} and comprises two main components: the input interface and the output analysis view.

The input interface is shown in Figure~\ref{fig:input-interface}. A toggle at the top of the interface determines whether the full article is provided as structured data (Figure~\ref{fig:input-interface}a) or as a URL (Figure~\ref{fig:input-interface}b). Once the user fills out the input data, they can start the analysis by hitting the 'Submit' button.

The output analysis view, which appears once the analysis is completed, is shown in Figure~\ref{fig:result}. The output from the model is broken down into three topical boxes. The top box shows the overall assessment; the lower two boxes cover the Location and Source, and Tampering.

The follow-up chat interface with the LLM is located below the output analysis viewer. The user can submit their questions using the field and button at the bottom; the messages cannot be deleted.

\section{Limitations}

While our proposed method and prototype address many challenges of previous approaches, both they and the analysis in this paper still have limitations, outlined in this section.

As LLMs are stochastic and, to a large extent, still not fully understood, undesired behaviors such as hallucination may yield incorrect results. In particular, LLMs lack a notion of truthfulness and often present information with confidence, even when it is incorrect or based on unknown sources~\cite{xu2024hallucination}. This could result in cases where the LLM used in our system lacks sufficient information to make a conclusive assessment but still provides one, potentially misleading the user. As more understanding and interventions are developed, this risk will be mitigated.

Second, as provenance metadata adoption is only starting to roll out into services and operating systems, most news articles, blog posts, and social media posts today do not have them. This could result in cases where the system is unable to extract any provenance metadata and the LLM still tries to make assessments about the story, leading to poorly sourced conclusions. However, we expect that this issue will, too, become smaller in time as wide-spread adoption is underway.

Given the nascent stage of provenance metadata's development, no datasets currently exist containing specific news articles with this kind of data. This limitation prevented us from conducting any benchmark evaluations and highlights a gap in existing scholarship: the need for a news dataset focused on the provenance metadata of attached media.

Furthermore, our method does not reflect shortcomings of the article present solely in its body. For instance, if the article has a partisan angle, is highly emotional, or includes a clickbait headline, our method would not flag it so long as the media in it are contextually relevant. The literature contains an extensive body of work on methods to detect and combat such textual manipulation~\cite{liu2024emotion, abdali2024multi}.

Finally, LLMs have been found to exhibit biases toward certain demographic groups (e.g., underrepresented ethnic, religious, and gender groups) as well as broader concepts~\cite{gallegos2024bias,guo2024bias}. Consequently, the model’s performance is likely to vary depending on the topics and entities present in the input article. While bias mitigation remains an active area of research~\cite{raza2024mbias,deng2024promoting}, and some of these issues may be alleviated in future model versions, such biases should be further studied and proactively addressed, particularly if this system or its derivatives are used in consumer-facing applications.

% TODO: even an article that passes this test can be manipulative, deceptive

\section{Conclusion}

We introduced a method that analyzes the relevance of visual media in news articles using LLMs and provenance metadata. Additionally, we developed a prototype web interface to interface with our method using structured article data or URLs.

% Presenting this demo at the EMNLP conference will foster discussion about this problem and allow us to gather extensive feedback from the community. We plan structure and share the findings from the demo with the research community.

% This feedback will enable future work to address any shortcomings and further enhance the system's robustness. 

Future work should prioritize mitigating issues such as hallucinations, bias, and lack of provenance metadata. Other areas for productive follow-up work include addressing edge cases (e.g., multiple images with varying assessments) and improving accessibility. This work should also perform a rigorous quantitative and qualitative evaluation of the performance of existing LLMs on this problem. This will require collecting a novel dataset, as there are no datasets suitable for this evaluation at the time of writing.

% mention that this is relevant for blogs
% \cite{Ando2005}

%\section*{Acknowledgments}
%Anonymized

\bibliography{custom}

\appendix

\newpage

\section{Prompts}
\label{app:prompt}

\begin{tcolorbox}[colback=lightgreen, colframe=black, title=System Prompt]
    You are evaluating the relevance and credibility of images and videos attached to news stories. Below, you will be presented with:
    \\
    -~the article's title;
    
    -~the article's body;
    
    -~captions of the attached media (images/videos);
    
    -~and provenance metadata of the attached media (images/videos), indicating information such as the origin (location and time) of the media, authors of the media, edits made to them, etc.
    \\
    Analyze the following:
    
    1.~Determine whether the location and time (year, month) when the media (images/videos) were taken is relevant to the news story.
    
    2.~Determine whether there was any tampering with the media (Photoshop edits, AI generation, etc.); resizing is okay.
    
    3.~Based on the above, give one of the following assessments: \texttt{'RELEVANT'} or \texttt{'NOT RELEVANT'}.
    \\
    Return your answer as a JSON/dictionary with the format: 
    
    \begin{verbatim}
{
'1-relevant': True/False, 
'1-reason': str, 
'2-relevant': True/False, 
'2-reason': str, 
'3-assessment': 
    'RELEVANT' / 'NOT RELEVANT'
}
\end{verbatim}
\end{tcolorbox}

\begin{tcolorbox}[colback=lightgray, colframe=black, title=Inference Prompt]

-~title: \texttt{[data]}

-~body: \texttt{[data]}

-~image caption: \texttt{[data]}

-~provenance metadata: \texttt{[data]}

\end{tcolorbox}

\begin{tcolorbox}[colback=lightgray, colframe=black, title=Follow-up Chat Prompt]
Here is your current reasoning: \texttt{[data]}. Generate a verbose response to the following question, highlighting the importance of provenance metadata: \texttt{[data]}.
\end{tcolorbox}

% \textbf{The following is the system prompt presented to the LLM:}

\end{document}